\documentclass[journal]{IEEEtran}
\usepackage{tikz}
\usepackage{amsfonts}
\usepackage{times}
\usepackage{soul}
\usepackage{url}
\usepackage[hidelinks]{hyperref}
\usepackage[utf8]{inputenc}
\usepackage{graphicx}
\usepackage{amsmath}
\usepackage{amsthm}
\usepackage{booktabs}
\usepackage{multirow}
\usepackage{mathtools}
\usepackage{algorithmic}
\usepackage[numbers]{natbib}
\usepackage[linesnumbered, ruled,vlined]{algorithm2e}
\SetKwInput{KwInput}{Input}                % Set the Input
\SetKwInput{KwOutput}{Output}

\newtheorem{theorem}{Theorem}
\newtheorem{lemma}[theorem]{Lemma}

\newcommand\Tstrut{\rule{0pt}{2.3ex}}         % = `top' strut
\newcommand\Bstrut{\rule[-1.3ex]{0pt}{0pt}}   % = `bottom' strut
             
\begin{document}

\title{Efficient Spiking Neural Networks with\\ Radix Encoding\vspace{-2pt}}

\author{Zhehui~Wang,
        ~Xiaozhe~Gu,
        ~Rick~Siow~Mong~Goh,
        ~Joey~Tianyi~Zhou, 
        and~Tao~Luo
\IEEEcompsocitemizethanks{ \IEEEcompsocthanksitem 
This work was supported by the Singapore Government’s Research, Innovation and Enterprise 2020 Plan (Advanced Manufacturing and Engineering domain) under Grant A1687b0033 and A1892b0026, and A18A1b0045, and Joey Tianyi Zhou's SERC Central Research Fund (Use-inspired Basic Research) (Corresponding author: Tao Luo)\\
Z. Wang, R. Goh, J. Zhou, and T. Luo are with the Institute of High Performance Computing, Agency for Science, Technology and Research (A*STAR), Singapore.\protect\\
\indent X. Gu is with the Chinese University of Hong Kong\\
\indent E-mail:\{wang\_zhehui, gohsm, joey\_zhou,  luo\_tao\}@ihpc.a-star.edu.sg
}}

\maketitle

\begin{abstract}

Spiking neural networks (SNNs) have advantages in latency and energy efficiency over traditional artificial neural networks (ANNs) due to their event-driven computation mechanism and the replacement of energy-consuming weight multiplication with addition. However, to achieve high accuracy, it usually requires long spike trains to ensure accuracy, usually more than one thousand time steps. This offsets the computation efficiency brought by SNNs because a longer spike train means a larger number of operations and larger latency. In this paper, we propose a radix-encoded SNN, which has ultra-short spike trains. Specifically, it is able to use less than six time steps to achieve even higher accuracy than its traditional counterpart. We also develop a method to fit our radix encoding technique into the ANN-to-SNN conversion approach so that we can train radix-encoded SNNs more efficiently on mature platforms and hardware. 
{Experiments show that our radix encoding can achieve 25X improvement in latency and 1.7\% improvement in accuracy compared to the state-of-the-art method, using the VGG-16 network on the CIFAR-10 dataset.} 
%works in various models, including VGG, ResNet, and MobileNet on datasets CIFAR-10, CIFAR-100, and ImageNet.
\end{abstract}

%ompared with the traditional encoding method, our radix encoding technique can achieve at least 167X speedup with even higher accuracy.


\begin{figure*}[!t]
  \centering
  \includegraphics[width=7in ]{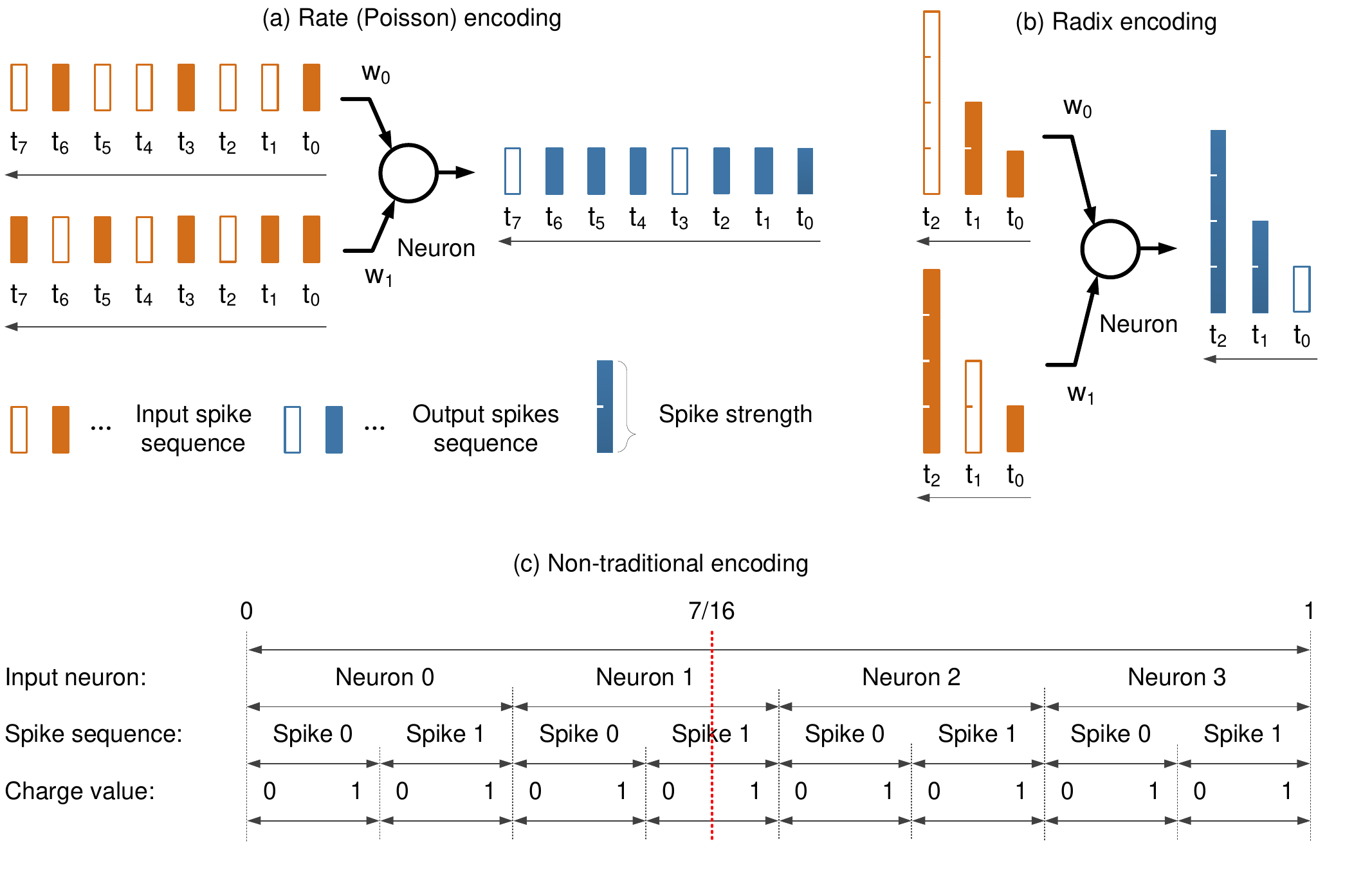}\\
  \caption{Comparison between the rate (Poisson) encoding, our proposed radix encoding and the non-traditional encoding scheme. The colored bars represent the existence of spikes while the hollow bars represent the absence of spikes, the height of bards represents the spike strength.}
  \label{f:compare_coding}
\end{figure*}

\section{Introduction}

\IEEEPARstart{D}{eep} neural networks, especially deep convolutional neural networks (CNNs), have been demonstrated as a prevailing technique in many complex real-world applications including computer vision (CV), natural language processing (NLP), and so on~\cite{krizhevsky2012imagenet}\cite{shen2016attention}\cite{ren2015faster}.
However, CNNs involve a large amount of weight and multiply-accumulate (MAC) operations, resulting in considerable energy consumption in data movement and MAC computations for model training and deployment.
The growing demand for edge AI and Artificial intelligence of things (AIoT) is exacerbating the situation, resulting in the emergence of efficient neural network design and accelerators, especially for inference~\cite{yang2017designing}\cite{sodhro2019artificial}.

The spiking neural networks (SNN) is a promising technology that has increasingly attracted people's attention in recent years. Theoretically, SNN has a higher similarity with the biological neural system.
It shows higher energy efficiency and lower latency than traditional artificial neural networks (ANNs) due to its special computation mechanism. The inference of SNN consists of only addition, which utilizes much fewer hardware resources than the complicated, energy-consuming multiplication operations used by ANN. Today, state-of-the-art works have pushed the accuracy of SNN to a competitive level with ANN. For example, Rueckauer's work shows over 90\% accuracy under the CIFAR-10 dataset~\cite{rueckauer2017conversion}. In the past decade, researchers not only advanced the model accuracy but also proposed many types of neuromorphic computing hardware systems~\cite{park201965}\cite{nambiar20200}\cite{luo2018fpga}\cite{pei2019towards}\cite{wang2020ncpower} to further improve the computation efficiency of SNN, such as SpiNNaker~\cite{khan2008spinnaker}, TrueNorth~\cite{merolla2014million}, and Loihi~\cite{davies2018loihi}.

However, in order to ensure high accuracy performance, traditional rate-encoded SNN models demand long spiking trains. A spike train can be considered as a sequence of events (binary) across multiple time steps. The length of the spike train is defined as the number of time steps. At each time step, the neuron may choose to fire a spike or not. In traditional rate-encoded design~\cite{diehl2015fast}, the signal strength in SNN is represented by the number of spikes over the number of time steps (i.e., the length of the spike train). Therefore, the precision of the signal strength is proportional to the length of the spike train. For example, ten-time-step corresponds to 10\% error, and one-hundred-time-step corresponds to 1\% error. To pursue high model accuracy, the signal strength expressed in SNN should be as precise as possible. Therefore, a long spike train becomes necessary for the rate-encoded SNN. Experiments show that to achieve competitive accuracy with the ANN, we need around one thousand time steps to encode the signal strength from the neuron~\cite{yan2021near}, resulting in low energy efficiency and longer inference time.

Unfortunately, long spike trains offset the advantages of SNN over ANN on computation efficiency. At each time step, neurons need to accumulate all the spikes from the input source and decide whether it should fire or not. More time steps mean more additional operations. Today, many works try to shorten the spike train~\cite{wu2019direct}\cite{zhang2019tdsnn}. However, it is difficult to do so without obvious accuracy loss. To solve this issue, we proposed the radix encoding method, where we assign different weights to spikes at different time steps. In other words, some spikes are more important than others when representing the signal strength. This is different from the traditional rate encoding method, where all the spikes have the same weight.
{The radix-encoding is a category of encoding schemes with different choices on assigned weights. In this paper, we mainly focus on the base-2 radix encoding. Specifically, if we use bit $1$ and bit $0$ to represent the existence and absence of spikes, the bit sequence of the spike train denotes the binary value of the data.}
With only a few time steps, we can represent the high-precision data. We mathematically verified the effectiveness of radix-encoded SNN on image classification. It shows much higher energy efficiency and much lower latency than the rate-encoded SNN.

We developed a method to fit our radix-encoded method into the ANN-to-SNN conversion approach~\cite{rueckauer2017conversion}\cite{perez2013mapping} so that we can train the radix-encoded SNN models more efficiently on mature platforms and hardware, using networks and datasets designed for ANN. To verify the efficiency of our design, we compared our radix encoding with the state-of-the-art method on the accuracy, the number of operations, and latency. Overall, the contributions of this work can be concluded as:
\begin{itemize}
\item We propose a radix-encoded SNN with ultra-short spike trains. This paper takes the base-2 radix-encoding as an example and mathematically proves its efficiency over the traditional rate-encoding.
\item We propose a method to fit the radix encoding technique into the ANN-to-SNN conversion approach so that we can train radix-encoded SNNs more efficiently on mature platforms and accelerators.
\item {We achieved a 25X improvement in latency and 1.7\% improvement in accuracy compared to the state-of-the-art method, using the VGG-16 network on the CIFAR-10 dataset.}

\end{itemize}
%We propose a radix-encoded SNN with ultra-short spike trains. Compared with the state-of-the-art works, our new design shows lower latency and higher accuracy.

%From expeirment results, we can observe larger imrpovoemt of radix encoding.
% The rest of the paper is organized as follows. Section~\ref{sec:relwork} gives the related work. Section~\ref{sec:preliminary} includes preliminaries, Section~\ref{sec:method} introduces the details of our radix-encoded SNN. Section~\ref{sec:experiment} discusses the experiments. Section~\ref{sec:conclusion} concludes.

{We organize the rest of the paper as follows. Section II presents a literature review on the topic of SNN training. Section III introduces three encoding schemes for SNN. Section IV presents our radix encoding scheme. Section V shows the experiment results of radix encoding and compares it with the state-of-the-art. Finally, Section VI presents our conclusion.
}

\section{Related Work}
\label{sec:relwork}
 
The training of spiking neural networks can be classified into unsupervised and supervised training. One popular unsupervised SNN training method is the spike-timing-dependent plasticity (STDP)~\cite{hebb1949organization}.
{We increase the weight value if the associated input spike triggers an output spike within a predefined time window and decrease the weight value if the associated input spike does not correlate with the output spike.}
For supervised learning, although it demonstrates a competitive accuracy than unsupervised learning, the training process is difficult because of the non-differentiability of spike trains.
There are mainly two solutions to alleviate this problem. {The first solution is to approximate the spiking function as a differentiable function ~\cite{neftci2019surrogate}.} 
{For example, Huh~\emph{et~al.} proposed a new threshold function for SNN neurons~\cite{huh2018gradient}. The new function is smoother than the original Dirac-delta function and is differentiable.}
However, this type of solution has a high computation cost because of the complicated approximating functions implemented during training. Today, many of these works are limited to shallow networks on simple dataset~\cite{shrestha2018slayer}\cite{kaiser2020synaptic}\cite{bellec2018long}.
{The second solution is to first train ANN using the standard training method and then convert the well-trained ANN to event-based SNN~\cite{rueckauer2017conversion}.}
In the literature, there are other techniques that can also help us to improve the training efficiency of SNN models. {For example, we can use a genetic algorithm to optimize the SNN model. Schuman~\emph{et~al.} proposed the EONS (Evolutionary Optimization for Neuromorphic Systems) framework~\cite{schuman2020evolutionary}, which can find the optimal SNN model architecture and train the model. Another technique is called the liquid state machine~\cite{reynolds2019intelligent}, an SNN version of reservoir computing. It simplifies the SNN training process because there is no need to tune the weights of the reservoir itself.}

Regarding the ANN-to-SNN conversion approach, since the activations in ANN are real numbers, while activations in SNN are spikes, we need to encode the spikes to float number and then decode the float number back to spikes. {Theoretically, shorting the encoding spike train will decrease the precision of the data, thus lowering the model accuracy. Several works in the literature have improved SNN models' accuracy using short spike trains. Yan~\emph{et~al.} developed a lossless transfer method from the ANN model to the SNN model~\cite{yan2021near}. They first trained the ANN model with clamped and quantized activations and then converted the ANN model parameters to corresponding SNN parameters. Sengupta~\emph{et~al.} improved the model accuracy by tuning the threshold of neurons~\cite{sengupta2019going}. They sampled the spiking activities of each layer and then developed an algorithm to balance the neuron's threshold based on the sampled data. Rathi~\emph{et~al.} did spike-based backpropagation on SNN models converted from ANN models~\cite{rathi2020enabling}. They updated the weights using the surrogate gradient function. In their surrogate function, the gradient is proportional to the distance of two spikes on the time domain. Lee~\emph{et~al.} also did spike-based backpropagation from ANN converted SNN models~\cite{lee2020enabling}. They used the straight-through estimator (STE)~\cite{yin2019understanding} to approximate the gradient. The derivative also included a correctional term, which can compensate for the leaky effect in the membrane potential. Kim~\emph{et~al.} shortened the spikes train and found the sources of noises that caused significant accuracy loss~\cite{kim2018deep}. They repeat the same input spike multiple times to compensate for that noise. To improve the accuracy, Hu~\emph{et~al.} applied the residual network architecture on SNN models by developing a shortcut conversion method for residual SNN~\cite{hu2018spiking}.} Different from the previous work, our radix encoding method can encode a high {precision} float number into an ultra-short spike train. There is no need to repeat spike trains in our radix-encoded SNN, but we can still achieve higher accuracy.
{
The above methods work in the traditional SNN dimension, where the spike sequence represents the signal. 
}
We will compare them with our work in Section~\ref{sec:experiment}.

 %between ours and them
%This feature further improves the efficiency of our radix-encoded SNN.

\section{Preliminaries}
\label{sec:preliminary}
{Unlike ANN, whose input/output are real numbers, the input/output of SNN are sequences of spikes, called spike trains.} A neuron receives spikes from upstream neurons via many synapses and generates spikes. Each synapse can receive one spike train sequentially and independently. Figure~\ref{f:compare_coding} describes the major difference between the rate (Poisson) encoding, our proposed radix encoding and the non-traditional encoding scheme.{
The first two encoding schemes work on the traditional SNN dimension, the spike sequence.}
In the rate (Poisson) encoding, the data is encoded by the neuron's firing rate. All spikes, either in the input spike train or in the output spike train, are identical. Each spike only represents a very small portion of the signal strength. To precisely represent the strength of the signals, many spikes are demanded. In our radix encoding, different spikes have different spike strengths depending on their locations on the spike train. Because of this additional information contained in the spike, our radix-encoded SNNs have ultra-short spike trains. {In addition to them, the non-traditional encoding scheme~\cite{schuman2019non} encodes the input data on two extra dimensions (input neuron and charge value) in addition to the traditional dimension (spike sequence).}

For example, Figure~\ref{f:compare_coding}(a) shows the traditional rate-encoded SNN. There are eight time steps in the spike train, and each spike can only represent 1/8 signal strength. Figure~\ref{f:compare_coding}(b) shows our proposed radix-encoded SNN, which uses only three time steps to achieve the same precision. This is because each spike represents different signal strength. More specifically, spikes in the first time step $t_0$ has a signal strength 1/8, while the spike in the $t_n$ time step has a signal strength $2^n$/8. To guarantee the precision 1/8, eight time steps are required in the rate-encoded SNN, while only three time steps are required in the radix-encoded SNN.
{Figure~\ref{f:compare_coding}(c) shows the non-traditional encoding scheme, where the second neuron needs to fire twice to encode the data 7/16 (red vertical line). Its first spike has a full charge value, and the second spike has a half charge value.}

\begin{algorithm}[t]
%\DontPrintSemicolon
\KwInput{binary sequence $i_n[t]$, weight $w_n$, bias $b$}
\KwOutput{binary sequence $o[t]$}
  $v = b$  \tcp*{Initialization} %\\
  \For{t $\leftarrow$ $0, 1, 2, \dots $}
    {
        \For{n $\leftarrow$ $0$ \KwTo $N-1$}
          { $v$ $\leftarrow$ $v + i_n[t] * w_n$ \tcp*{Integration} 
          }
        \If{$\hat{v}$ $\geq$ $\Phi$}
               {$o[t] = 1$ \tcp*{Firing} 
               $v=v-\Phi$   \tcp*{Refractory Period}
               }
            \Else
            {
            $o[t] = 0$\;
            }
         
      $v$ $\leftarrow$ $v*\kappa$\tcp*{Leakage}

    }
\caption{Leaky integrate-and-fire (LIF) neuron model for radix-encoded SNN}
\label{neuron}
\end{algorithm}

\section{SNN with Radix Encoding}
\label{sec:method}
We can mathematically verify the effectiveness of radix-encoded SNN on image classification. First of all, a spike train can be expressed in Equation~(\ref{sequence}). Each element in the spike train stands for the existence of the spike. By definition, $s(t)$ equals $1$ if the spike exists at time step $t$, and equals $0$ if not. $T$ is the length of the spike train.
\begin{equation}
s(t) \in \{0, 1\}\ \ \ \ \ \ \  t= 0, 1, 2, ..., T-1
\label{sequence}
\end{equation}
{$T$ is an important parameter. An SNN model can have various choices of $T$. Theoretically, the change of $T$ has large impacts on the model accuracy and latency. In the experiment part, we will show the performance of radix-encoded SNNs under different $T$.}

%with Diffent lengths of spke train length have differnet accracy and letnwcy. 
% We denote the input sequence from $n$-th synapse as $i_n(t)$, and express it in Equation~(\ref{input}). Here $s_n^{in}(t)$ stands for the input spike train from $n$-th synapse. If the time step $t$ is greater than the spike train length $T$, $i_n(t)$ always equals $0$.
% \begin{equation}
% i_n(t)  = 
%  \begin{cases}
%  s_n^{in}(t)  & \text{if \ \ $0 \leq t < T$} \\
%  0 &  \text{if \ \  $t\geq T$} 
%  \end{cases}
% \label{input}
% \end{equation}

\begin{figure}[!t]
\begin{tikzpicture}

\draw(-0.76, 11.7) node{Input};
\draw(1.63, 11.7) node{Output};

\filldraw[black] (-0.25, 11.0) circle (2 pt);
\filldraw[black]  (-0.25, 10.0) circle (2 pt);
\filldraw[black]  (-0.25, 8.0) circle (2 pt);

\filldraw[black]  (1.25, 11.0) circle (2 pt);
\filldraw[black]  (1.25, 10.0) circle (2 pt);
\filldraw[black]  (1.25, 8.0) circle (2 pt);

\draw(-0.78, 11.0) node{$i_0(t)$};
\draw(-0.78, 10.0) node{$i_1(t)$};
\draw(-0.55, 7.6) node{$i_{N-1}(t)$};

\draw(1.80, 11.0) node{$o_0(t)$};
\draw(1.80, 10.0) node{$o_1(t)$};
\draw(1.55, 7.6) node{$o_{M-1}(t)$};

\draw[-] (-0.25,11.0) -- (1.25,11.0);
\draw[-] (-0.25,11.0) -- (1.25,10.0);
\draw[-] (-0.25,11.0) -- (1.25,8.0);

\draw[-] (-0.25,10.0) -- (1.25,11.0);
\draw[-] (-0.25,10.0) -- (1.25,10.0);
\draw[-] (-0.25,10.0) -- (1.25,8.0);

\draw[-] (-0.25,8.0) -- (1.25,11.0);
\draw[-] (-0.25,8.0) -- (1.25,10.0);
\draw[-] (-0.25,8.0) -- (1.25,8.0);

\node [rotate=90] at (-0.78, 8.8) {\LARGE...};
\node [rotate=90] at (1.80, 8.8) {\LARGE...};

\draw(0.5, 6.9) node{(a)};

\draw(5.25, 11.7) node{$m$-th Output Neuron};

\draw[-] (3.9,11.0) -- (4.9,11.0);
\draw[->, >=stealth] (3.9,10.0) -- (4.9,10.0);
\draw[-] (3.9,8.0) -- (4.9,8.0);

\draw[->, >=stealth] (4.9,11.0) -- (5.2,10.13);
\draw[->, >=stealth] (4.9,8.0) -- (5.2,8.87);

\draw[->, >=stealth] (5.92,9.5) -- (6.5,9.5);

\draw(3.4, 11.0) node{$i_0(t)$};
\draw(3.4, 10.0) node{$i_1(t)$};
\draw(3.635, 7.6) node{$i_{N-1}(t)$};

\draw(7.1, 9.5) node{$o_m(t)$};

\node [rotate=90] at (3.4, 8.8) {\LARGE...};

\draw (5.3,9.5) circle (18 pt);
\draw(5.3, 9.5) node{$v_m(t)$};

\draw(4.4, 11.25) node{$w_{0m}$};
\draw(4.4, 10.25) node{$w_{1m}$};
\draw(5.7, 8.0) node{$w_{N-1,m}$};

\draw(5.25, 6.9) node{(b)};
\draw (4.1,4.9) circle (18 pt);
\draw (6.0,4.9) circle (18 pt);

\draw (2.2,3) circle (18 pt);
\draw (4.1,3) circle (18 pt);
\draw (6.0,3) circle (18 pt);
%\draw (9,3) circle (20 pt);

\draw[->, >=stealth] (2.7,3.5) -- (3.6,4.4);
\draw[->, >=stealth] (4.6,3.5) -- (5.5,4.4);
\draw[->, >=stealth] (4.1,4.2) -- (4.1,3.7);
\draw[->, >=stealth] (6.0,4.2) -- (6.0,3.7);

\draw(2.2, 3) node{$v_m(-1)$};
\draw(4.1, 3) node{$v_m(0)$};
\draw(6.0, 3) node{$v_m(1)$};

\draw(4.1, 4.9) node{$v^{\prime}_m(0)$};
\draw(6.0, 4.9) node{$v^{\prime}_m(1)$};

\draw(-0.31, 4.9) node{Integration:};

\draw(0, 3) node{Firing/Leakage:};

\draw(2.2, 6) node{t = -1};
\draw(4.1, 6) node{t = 0};
\draw(6.0, 6) node{t = 1};

\draw(7.3, 4.9) node{\LARGE...};
\draw(7.3, 3) node{\LARGE...};

\draw(3.25, 1.85) node{(c)};

\end{tikzpicture}
\caption{{(a) A simplified two-layer model with N input neurons and M output neurons; (b) The details of the $m-$th output neuron; (c) The transitions of membrane potential from time step $-1$ to time step $1$.}} 
\label{f:r1}
\end{figure}

\subsection{Working Principles of SNN Neurons}

The operation mechanism of neurons in the radix-encoded SNN is shown in Algorithm~\ref{neuron}. Each neuron has an internal parameter called the membrane potential. In the beginning, this value is initialized to $b$ (line 1). Next, the value of the membrane potential is updated every time step. The neuron behavior at time step $t$ is based on the value of membrane potential at the previous time step $t-1$. In the leaky integrate-and-fire (LIF) model~\cite{hunsberger2015spiking}, we can divide a single time step into three stages. The first stage is called integration. The neuron receives spikes from all the synapses and accumulates these spikes with different weights into the neuron's membrane potential $v$ (lines 3-4). The second stage is called firing. If the membrane potential meets a certain condition, the neuron fires a spike via the axon (lines 5-6). If the spike is fired, the membrane potential will go through the refractory period (line 7). Here $\hat{v}$ denotes the least significant digit of $v$ in the encoded format, and $\Phi$ is the threshold. The third stage is called leakage. The membrane value would decrease by $\kappa$ (line 10). Here $\kappa$ is the leak factor.

%$o_m(t)$ is the output sequence, which can be either $0$ or $1$.

{Figure~\ref{f:r1}(a) shows a simplified two-layer SNN model with $N$ input and $M$ output neurons. The signal from each SNN neuron is a spike train, which is a function of the time step index $t$. We use notations $i_n(t)\ (0\leq n< N)$ to denote signals from input neurons and notations $o_m(t)\ (0\leq m< M)$ to denote signals from output neurons, respectively. Figure~\ref{f:r1}(b) shows the details of the $m-$th output neuron. We denote the weight associated with the $n-$th input neuron by $w_{nm}$. The membrane potential of this neuron is also a function of the time step index $t$. In this paper, we use $v_m(t)$ to denote the membrane potential of the $m-$th output neuron after performing all the operations of time step $t$.}

{According to Algorithm~\ref{neuron}, each time step has three stages: the integration stage, the firing stage, and the leakage stage. To study the membrane potential changes after each stage, we use different notations for different stages. Figure~\ref{f:r1}(c) shows the transitions of the $m-$th output neuron. Let $v^{\prime}_m(t)$ denote the membrane potential after the integration stage, and $v_m(t)$ denote the potential after both the firing stage and the leakage stages. Before step $0$, the neuron is in its default state with membrane potential $v_m(-1)$. Afterward, the transition of the membrane potential follows the arrows in the figure. For example, at time step $0$, the membrane potential changes from $v_m(-1)$ to $v^{\prime}_m(0)$, and then changes to $v_m(0)$.}

\begin{figure}[!t]
  \centering
  \includegraphics[width=3.5in]{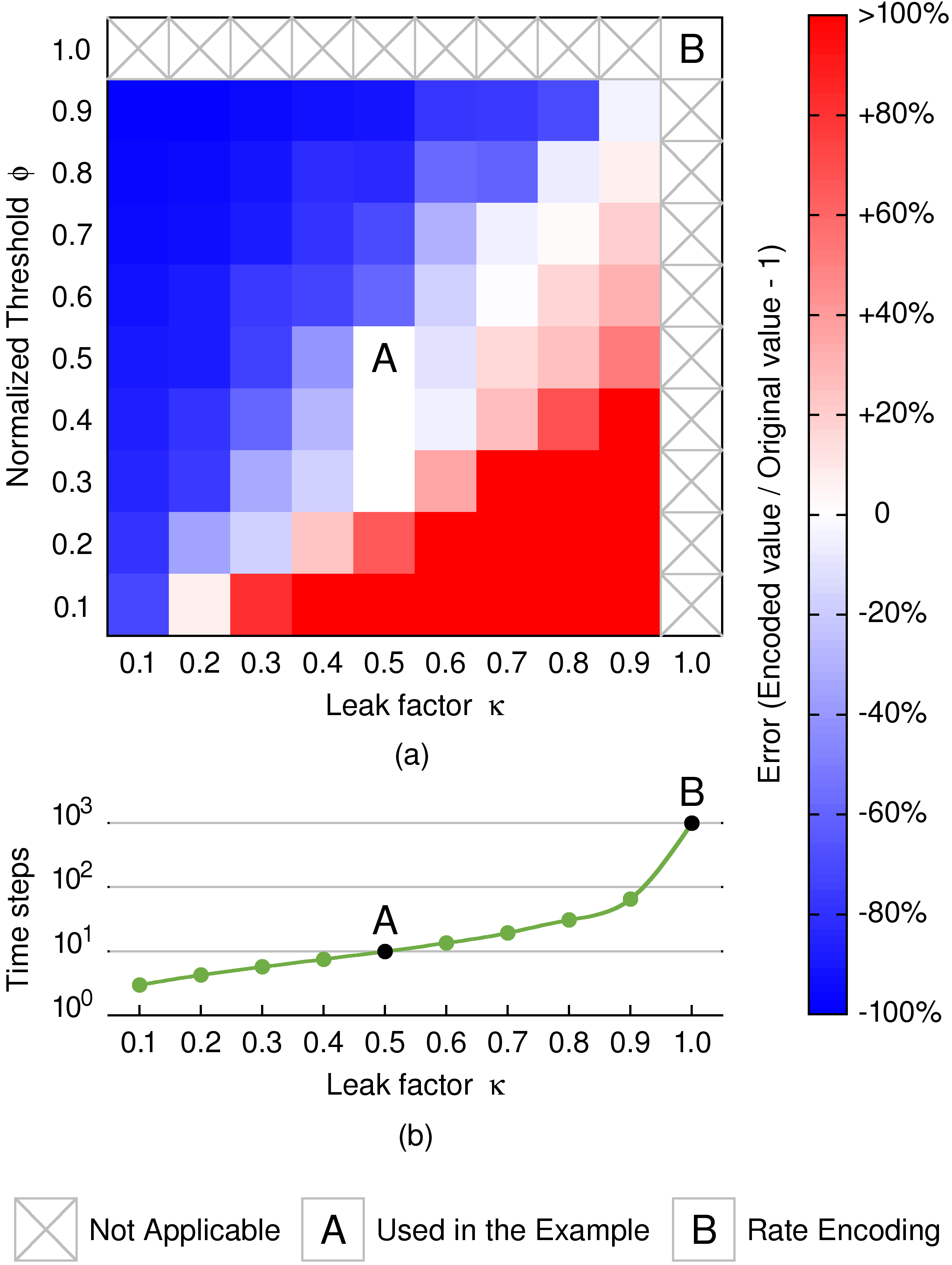}\\
  \caption{{(a) The error of radix encoding under various leak factors $\kappa$ and normalized thresholds $\phi$. We randomly pick up one million real numbers from $0$ to $1000$ and find the average encoding error; (b) The relationship between the number of time steps and the leak factor.}}
  \label{f:r2}
\end{figure}

\begin{figure*}[!t]
  \centering
  \includegraphics[width=7in ]{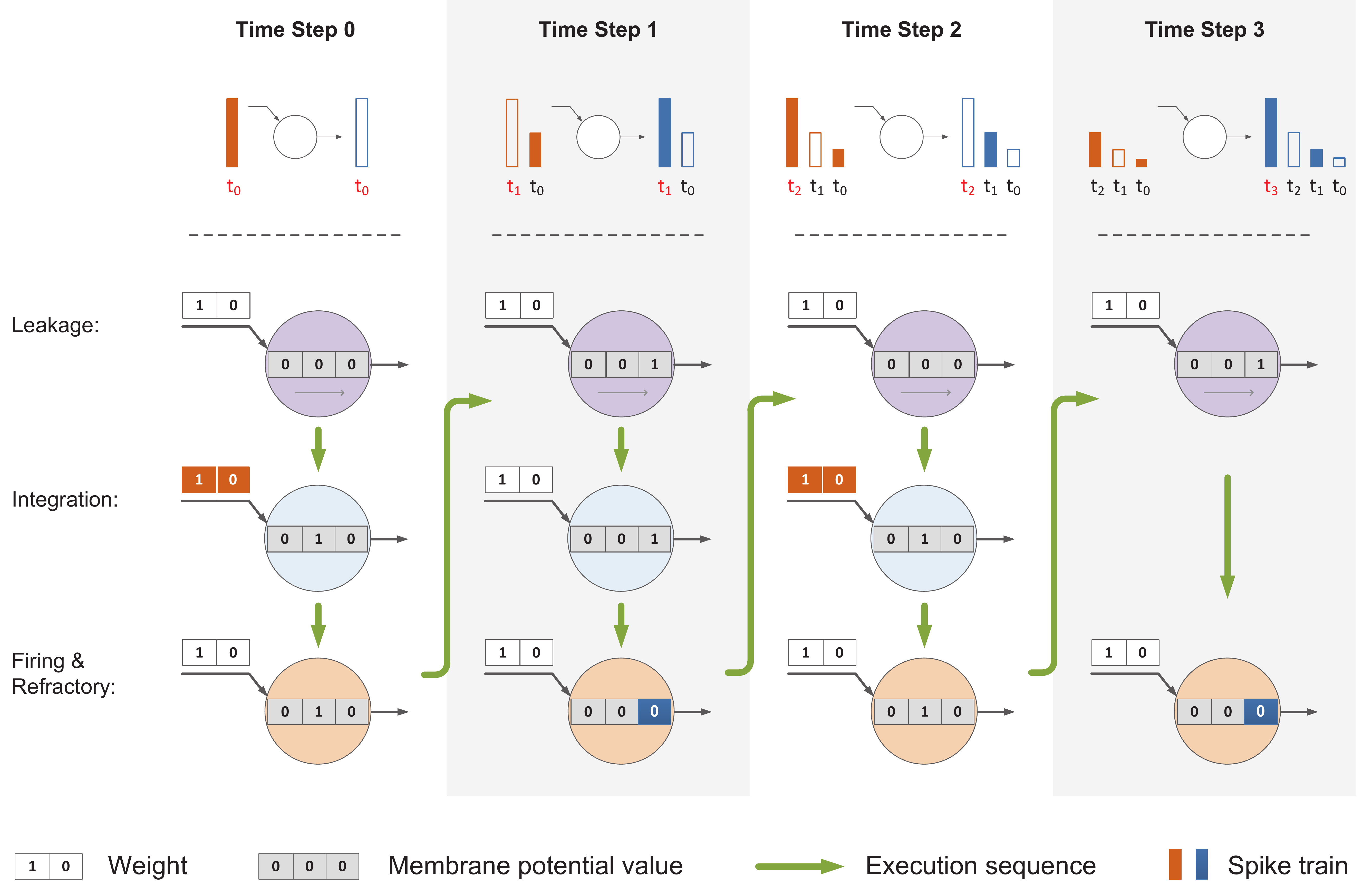}\\
  \caption{An example of radix-encoded SNN in the one-input neuron. At the top of the figure, the colored bars represent the existence of the spikes while the hollow bars represent the absence. The height of the bars represent the spike strength.}
  \label{f:process_path}
\end{figure*}

We can derive $v_m(t)$, the membrane potential at time step $t$ from Algorithm~\ref{neuron}. At the very beginning, the membrane potential $v_m(-1)$ is initialized by a bias value $b_m$. We expressed it in Equation~(\ref{bias}).
\begin{equation}
v_m(-1) = b_m
\label{bias}
\end{equation}
After passing the integration stage, the membrane potential changes from $v_m(t-1)$ to $v^{\prime}_m(t)$. We can express the relationship between $v_m(t-1)$ and $v^{\prime}_m(t)$ in Equation~(\ref{before}). 
\begin{equation}
v^{\prime}_m(t) = v_m(t-1) + \sum_{n=0}^{N-1}{i_n(t)\cdot w_{nm}} 
\label{before}
\end{equation}
After passing the firing stage and the leakage stage, the membrane potential changes from $v^{\prime}_m(t)$ to $v_m(t)$. We can express the relationship between $v^{\prime}_m(t)$ and $v_m(t)$ in Equation~(\ref{after}). 
\begin{equation}
v_m(t) =\kappa \cdot(v^{\prime}_m(t)-\Phi\cdot o_m(t))
\label{after}
\end{equation}

\subsection{Exploration of Hyper-Parameters}

Radix encoding is compatible with a wide range of leak factors $\kappa$ and thresholds $\Phi$. Theoretically, the leak factor $\kappa$ can range from $0$ to $1$, and the threshold $\Phi$ can range from $0$ to $1/\kappa$, where $1/\kappa$ is the base of the coding scheme. We normalize the threshold to be $\phi = \Phi\cdot \kappa$. Thus, both hyperparameters $\kappa$ and $\phi$ range from $0$ to $1$. Changing the values of hyperparameters also changes the sequence of encoding spikes, resulting in different encoding errors.
We use the error metric defined in Equation~(\ref{e:error_ref}) to evaluate the difference between the encoded data value and the original data value.
{
In the equation, $E_{\kappa,\phi}$ denotes the encoded error under hyper-parameters $\kappa$ and $\phi$. The term $x_o$ is the original data value and $x_e$ is the encoded data value. $s(t)$ is the sequence of the spikes that encode data $x_o$.
}
The error can range from $-100\%$ to positive infinity. If the encoded value is close to the original value, the error is near $0$. If the encoded value is far from the original value, we can observe either positive or negative error with a large absolute value.
\begin{equation}
E_{\kappa, \phi}(x_o)= \frac{x_e}{x_o} -1 = \sum_t{(s(t)\cdot\phi \cdot  \kappa^{-t-1})}\cdot x_o^{-1}-1
\label{e:error_ref}
\end{equation}

For example, let us assume the original data value $x_o$ to be $200$. If the hyper-parameters $\kappa$ and $\phi$ are $0.5$ and $0.5$ respectively (binary encoded), the sequence of the spikes $s(t)$ is `11001000' (according to Algorithm~\ref{neuron}), where digit $1$ and $0$ represent the existence and absence of spikes. Refer to Equation~(\ref{e:error_ref}), the encoded data value $x_e$ equals $200$, the same as the original data value $x_o$. Hence, the encoded error $E_{\kappa=0.5, \phi=0.5}(200)=0$. Alternatively, if we set the hyper-parameters $\kappa$ and $\phi$ to be $0.2$ and $0.2$ respectively, the sequence of the spike $s(t)$ is `1100'. In that case, the encoded data value $x_e$ equals $150$, which is $25\%$ smaller than the original data value $x_o$. In other words, the encoded error $E_{\kappa=0.2, \phi=0.2}(200)=-0.25$.

% where 1 represents the existance of spikes and 0 represent the absence of spikes. the econded value  is 12**3 +02**2+ 0*2**1+ 1= 5 , then the error is 10, since the ecnode dataa eqila the orginal value, we have no erroos at all.In other case, if the normalzued threrhold and the leak factor is 0.2 and 0.1 respectvely, to encodie the data 10, the spike is 1000, orogual value is 20,1 and the encoded value is 011, 0.5*0.5*5**2 then the error is 20\

We use the Monte Carlo method~\cite{metropolis1949monte} to evaluate the encoding error under different hyper-parameters. Specifically, we randomly pick up one million real numbers from $0$ to $1000$ and find the average encoding error. Figure~\ref{f:r2}(a) shows the errors of the radix encoding under various leak factors $k$ and normalized thresholds $\phi$. The red and blue blocks denote the positive and negative errors in the heat map, respectively. The white blocks represent errors with tiny absolute values. Figure~\ref{f:r2}(b) shows the relationship between the number of time steps (in log scale) and the leak factor $\kappa$. As we can see from the figure, the number of time steps increases exponentially if the leak factor increases.

{From Figure~\ref{f:r2}(a), we can observe a long region containing several white blocks. The hyperparameters associated with those blocks are good candidates for achieving high model accuracies because of their lower encoding errors. From Figure~\ref{f:r2}(b), we can conclude that a smaller leak factor is preferred because it has fewer time steps. Therefore, considering the trade-off between the model accuracy and the number of time steps, we set $\kappa$ and $\phi$ to be $0.5$ and $0.5$ (base-$2$), respectively (point A). We take this setting as the example case. Note that the rate encoding is also a particular case in our radix encoding framework~\cite{yan2021near}, whose leak factor and threshold equal $1$ and $1$, respectively (point B). As we can see from Figure~\ref{f:r2}(b), this setting uses significantly more time steps.
}

\subsection{Base-2 Radix-Encoded SNN Model}

We use the selected setting (base-$2$: $\kappa = 0.5$ and $\phi = 0.5$) to build a radix-encoded SNN model. Based on Equation~(\ref{bias}) to~(\ref{after}), we can derive $v_m(t)$ in Equation~(\ref{inference}). 
\begin{equation}
\begin{aligned}
v_m(t) &= \sum_{n=0}^{N-1}{\sum_{\tau=0}^{t}{2^{-\tau-1}} i_n(t-\tau)\cdot w_{nm}} + 2^{-t-1} b_m \\ &-\sum_{\tau=0}^{t}{2^{-\tau-1}}o_m(t-\tau)
\end{aligned}
\label{inference}
\end{equation}

In Figure~\ref{f:process_path}, we show how the base-$2$ radix-encoded SNN works. For simplicity reasons, the neuron in the example has only one input. We assume the weight associated with this input is a $2$-bit integer `10', which equals $2$ in decimal format. The input spike train has three time steps, and the spikes exist in the first and third time step. This spike train represents the binary value `101', which equals $5$ in decimal format. The whole execution process contains four time steps. Each time step has three stages: the leakage stage, the integration stage, and the firing/refractory stage. Noted that in Algorithm~\ref{neuron}, the leakage stage is executed after the other two stages in every time step. In this example, for alignment reasons, the leakage stage is executed before the other two stages. There is no difference in the computation mechanism by shifting the position of the leakage stage from the end of each time step to the beginning. This is because the first stage of one time step can be considered as the last stage of the previous time step. We also show the status of the input/output spike trains of each time step at the top of the figure. The colored bars represent the existence of the spikes, while the hollow bars represent the absence. The height of the bars represents the spike strength. 

In this example, the neuron takes four time steps to complete the whole computation process:

\begin{itemize}

\item Time Step $0$: The value of the membrane potential is reset to be $0$, shown as `000' in binary format. After the leakage stage, its value is still `000'. The next stage is integration. Because of the existence of the input spike, the weight value is added to the membrane potential, which becomes `010'. Since the last digit of the membrane potential value is $0$, the neuron does not meet the firing condition at this time step. There is also no need to go through the refractory period.

% the firing condition does not been met at this time step. 

\item Time Step $1$: After the leakage stage, the value of the membrane potential is scaled by the leak factor $0.5$. We implement this by shifting all binary digits right by one bit, changing the value from `010' to `001'. In this time step, due to the absence of an input spike, nothing happens in the integration stage. However, since the last digit of the membrane potential value now becomes `1', there is one spike fired in this time step. The refractory period is also activated, which turns the value of the membrane potential back to `000'.

\item  Time Step $2$: It takes similar procedures to process the data, and the membrane potential becomes `010', which is greater than $0$. Therefore, another time step is necessary because more spikes may be fired later. 

\item Time Step $3$: In this step, there is no integrating stage since the input spike train has only three time steps. In the firing/refractory stage, the neuron fires a spike, and the membrane potential goes back to `000'. As it is less than $1$, no spikes would be fired later, so the whole procedure ends.

\end{itemize}

% As it is less than $1$, no spikes would be fired later, so the whole procedure stops.

After the leak leakage of every time step, the value of the membrane potential is reduced by half. Therefore, the new spikes sent into or fired out from the neuron at this time step have relatively doubled spike strength compared to previous spikes. In the figure, to clearly show this relationship, we reduce the height of all previous spikes by half after every leakage stage. Mathematically, the operation in the radix-encoded SNN is equivalent to the multiplication operation in ANN. We mathematically prove this in the next part (refer to Equation~(\ref{approx})). In this case, the output spike train represents binary value `1010', or $10$ in decimal format, which equals the product of weight (equaling $2$) and input spikes (equaling $5$). If the length of the output spike train is greater than the length of the input spike train, we can align them by removing the first few spikes of the output spike train because these spikes are less significant in terms of strength. Back to our example, we can see that the input spike train has three time steps while the output spike train has four times steps. Hence, we only keep the last three time steps of the output spike train, changing the output spike train from `1010' to `101'. This is equivalent to scaling the input spike train by the factor $2^{-\Delta T}$ ($\Delta T=1$). We mathematically explain this process in 
the next part (refer to Equation~(\ref{new_new_relationship})).

\subsection{Mathematical Proof of Radix Encoding}
We prove the effectiveness of radix-encoded SNN by mathematically finding its equivalence to ANN model. Based on the previous definition, we have Lemma~\ref{lemma1}.

\begin{lemma}
If $i_n(t) = 0$ for any $t\geq T$, There exists an integer $T^{\prime}>T$, such that when $t\geq T^{\prime}-1$, series $v_m(t)$ (refer to Equation~(\ref{inference})) converges to constant $c$, where $c = 0$ or $c = -1$.
\label{lemma1}
\end{lemma}

\begin{proof} 
Since $T^{\prime}>T$, we have $t>T-1$, the relationship between the membrane potential at time step $t+1$ and time step $t$ can be expressed in Equation~(\ref{same_side}).
\begin{equation}
v_m(t+1) = 2^{-1}(v_m(t)-o_m(t))
\label{same_side}
\end{equation}
We denote $\Delta v_m(t)$ as the difference between $v_m(t+1)$ and $v_m(t)$, and express it in Equation~(\ref{delta})
\begin{equation}
\Delta v_m(t) = v_m(t+1) - v_m(t) =  -2^{-1}(v_m(t)+o_m(t))
\label{delta}
\end{equation}
Since $i_n(t)$ is a binary sequence, $w_{nm}$ and $b_m$ are integers, $v_m(T-1)$ must be an integer. Its value can be either greater than or equals to $0$, or it can be less than or equals to $-1$. We discuss these two possibilities one by one. If $v_m(T-1)$ is greater than or equals to $0$, we can infer from Equation~(\ref{same_side}) that $v_m(t)$ is always greater than or equals to $0$, for any $t$ greater than $T-1$. On the other hand, we can infer from Equation~(\ref{delta}) that $\Delta v_m(t)$ is less than or equals to $-1$, if $v_m(t)$ is greater than $0$. We list these two conclusions in Equation~(\ref{con_a1}) and~(\ref{con_a2}). Combine them together, it can be proved that the series $v_m(t)$ will eventually converge to $0$.
\begin{align}
 \label{con_a1}
 &v_m(t)\geq 0  & \ &\text{if $v_m(T-1)\geq 0$ \ \ } \\
 &\Delta v_m(t) \leq -1  & \  &\text{if $v_m(t)> 0$ } 
 \label{con_a2}
\end{align}
If $v_m(T-1)$ is less than or equals to $-1$, based on the similar logic, we can have two conclusions, listed in Equation~(\ref{con_b1}) and~(\ref{con_b2}). Combine them together, it can be proved that the series $v_m(t)$ will eventually converge to $-1$.
\begin{align}
\label{con_b1}
&v_m(t) \leq -1 & \ & \text{if $v_m(T-1)\leq -1$} \\
&\Delta v_m(t) \geq 1 \ \ \ & \ & \text{if $v_m(t) < -1$ } 
\label{con_b2}
\end{align}

\end{proof} 

%Here $s_n^{in}(t)$ stands for the 
%Noted that $T^{\prime}$ is greater than $T$, for alignment reasons,

From Lemma~\ref{lemma1}, we show that the sequence of $v_m(t)$ would finally converge to either $0$ or $-1$, when $t = T^{\prime}-1$. We assume that the input sequence $i_n(t)$, weight matrix $w_{nm}$ and bias $b_m$ meet the condition~(\ref{condition}). The following derivations are all based on this condition.
\begin{equation}
\sum_{n=0}^{N-1}{\sum_{\tau=0}^{T^{\prime}-1}{2^{-\tau-1}} i_n(T^{\prime}-\tau-1)\cdot w_{nm}} + 2^{-T^{\prime}} b_m\geq 0
\label{condition}
\end{equation}
From Equation~(\ref{inference}) we can infer that $v_m(T^{\prime}-1) > -1$ under the condition~(\ref{condition}). This is because of the following Inequality~(\ref{inference_help}). 
\begin{equation}
\sum_{\tau=0}^{T^{\prime}-1}{2^{-\tau-1}}o_m(T^{\prime}-\tau-1) < 1
\label{inference_help}
\end{equation}
Since $v_m(T^{\prime}-1)$ can be either $0$ or $-1$, we have  $v_m(T^{\prime}-1)=0$, under the condition~(\ref{condition}). Put it back to Equation ~(\ref{inference}), the relationship between the input sequence $i_n(t)$ and the output sequence $o_m(t)$ can be expressed in Equation~(\ref{relationship}).
\begin{equation}
\begin{multlined}
 \sum_{\tau=0}^{T^{\prime}-1}{2^{-\tau-1}}o_m(T^{\prime}-\tau-1) =\\ \sum_{n=0}^{N-1}{\sum_{\tau=0}^{T^{\prime}-1}{2^{-\tau-1}} i_n(T^{\prime}-\tau-1)\cdot w_{nm}} + 2^{-T^{\prime}} b_m 
\label{relationship}
\end{multlined}
\end{equation}
By scaling Equation~(\ref{relationship}) on both sides with the factor $2^{T^{\prime}}$, and substituting $T^{\prime}-\tau-1$ with $t$, we have Equation~(\ref{new_relationship}).
\begin{equation}
\sum_{t=0}^{T^{\prime}-1}{2^{t}}o_m(t) = \sum_{n=0}^{N-1}{\sum_{t=0}^{T^{\prime}-1}{2^{t}} i_n(t)\cdot w_{nm}} + b_m 
\label{new_relationship}
\end{equation}

In our previously introduced radix-encoded SNN (refer to Part C of this section), the spike train $i_n(t)$ and $o_m(t)$ contain several insignificant spikes, which have little impact on data precision. Hence, we need to extract those significant spikes from $i_n(t)$ and $o_m(t)$ and align them. We denote the extracted spike train from the $n$-th input neuron as $s_n^{in}(t)$, and the extracted spike train from the $m$-th output neuron as $s_m^{out}(t)$. They are expressed in Equation~(\ref{input}) and (\ref{output}). $s_n^{in}(t)$ and $s_m^{out}(t)$ have the same spike train length $T$.
\begin{align}
\label{input}
s_n^{in}(t) &= i_n(t) &0\leq t<T\\
\label{output}
s_m^{out}(t) &= o_m(t+T^{\prime}-T) &0\leq t<T
\end{align}
By putting Equation~(\ref{input}) and~(\ref{output}) back to Equation~(\ref{new_relationship}), substituting $T^{\prime}-T$ by $\Delta T$, and scaling Equation~(\ref{new_relationship}) on both sides with $2^{-\Delta T}$, we have Equation~(\ref{new_new_relationship}). 
\begin{equation}
\begin{multlined}
\sum_{t=0}^{T-1}{2^{t}}s_m^{out}(t) + \sum_{t=0}^{\Delta T-1}{2^{t-\Delta T}}o_m(t) =\ \ \ \ \ \ \ \ \ \ \ \ \ \ \ \ \ \ \\
\sum_{n=0}^{N-1}{\sum_{t=0}^{T-1}{2^{t}} s_n^{in}(t)\cdot2^{-\Delta T} w_{nm}} + 2^{-\Delta T} b_m
\label{new_new_relationship}
\end{multlined}
\end{equation}

\renewcommand{\arraystretch}{1.6}
\begin{table}[!t]
\caption{Data and Model Transformation Functions}
\resizebox{1.00\linewidth}{!}{
\begin{tabular}{c | c }
\hline
Input $x, X$& Data Transform \& Inverse Transform \Bstrut\\
\hline
$ s_n^{in}(t)$, $ s_m^{out}(t)$   & $\textrm{WDT}(x) =\sum\limits_{t=0}^{T-1}{2^{t}}x(t)$ \Tstrut \Bstrut\\
$ S_n^{in}$, $ S_m^{out}$    & $\textrm{WDT}^{-1}(X) =\lfloor X/ {2^t}\rfloor ~\textrm{mod}~2 $ \Tstrut\Bstrut\\
\hline
Input $x, X$ & Parameter Transform \& Inverse Transform \Bstrut\\
\hline
$ w_{nm}, b_m $ & $\textrm{WPT}(x) =2^{-\Delta T}x $ \Tstrut \Bstrut\\
$ W_{nm}, B_m $ & $\textrm{WPT}^{-1}(X) =2^{\Delta T}X$\Tstrut\Bstrut\\
\hline
\end{tabular}
}
\label{transform}
\end{table}

To make the Equation~(\ref{new_new_relationship}) easy to read, we define some new notations for those complicated terms, shown in Equations~(\ref{new_denote}). Noted that $s_n^{in}(t)$ and $s_m^{out}(t)$ are binary sequences. Their elements can be either $0$ or $1$. The capitalized $S_n^{in}$ or $S_n^{in}$ is a single real number. The weight $w_{nm}$ and bias $b_m$ are integers, while the capitalized weight $W_{nm}$ and bias $B_m$ can be any real number.
\begin{equation}
\begin{aligned}
&S_n^{in} = \sum_{t=0}^{T-1}{2^{t}}s_n^{in}(t) &\ \ \  &S_m^{out} =  \sum_{t=0}^{T-1}{2^{t}}s_m^{out}(t) \\
&W_{nm}= 2^{-\Delta T}w_{nm}  &\ \ \  & B_{m}= 2^{-\Delta T}b_{m} 
\end{aligned}
\label{new_denote}
\end{equation}
By putting the new notations listed in Equations~(\ref{new_denote}) back to Equation~(\ref{new_new_relationship}), we have Equation~(\ref{simplified}).
\begin{equation}
S_m^{out} + r_m = \sum_{n=0}^{N-1}{S_n^{in}\cdot W_{nm}} + B_m 
\label{simplified}
\end{equation}
Here, the residue $r_m$ is related to the first few elements in the sequence $o_m(t)$, which can be expressed in Equation~(\ref{residue}). It is easy to prove that $r_m$ is less than $1$.
\begin{equation}
r_m = \sum_{t=0}^{\Delta T-1}{2^{t-\Delta T}}o_m(t) <1
\label{residue}
\end{equation}
Since $S_m^{out}$ is on the scale of $2^T$, which is much larger than $r_m$, we do some approximations, and rewrite Equation~(\ref{simplified}) as approximate Equation~(\ref{approx}).
\begin{equation}
S_m^{out} \approx  \sum_{n=0}^{N-1}{S_n^{in}\cdot W_{nm}} + B_m
\label{approx}
\end{equation}
Starting from Equation~(\ref{relationship}), our derivations are all based on the condition~(\ref{condition}). By putting notations in Equations~(\ref{new_denote}) back to the condition~(\ref{condition}), we have the condition~(\ref{condition2}).
\begin{equation}
\sum_{n=0}^{N-1}{S_n^{in}\cdot W_{nm}} + B_m \geq 0
\label{condition2}
\end{equation}

\begin{figure}[!t]
\begin{tikzpicture}
\draw (1,3) circle (29 pt);
\draw (7.0,3) circle (29 pt);
\draw(4.0,3.5) node{$W_{nm}= \textrm{WPT}(w_{nm}$)};
\draw(4.0,2.5) node{$B_{m}= \textrm{WPT}(b_{m})$};
\draw(4.0,0.5) node{$S_m^{out} = \textrm{WDT}(s_m^{out}(t))$};
\draw(4.0,5.5) node{$S_n^{in} =  \textrm{WDT}(s_n^{in}(t))$};
\draw(1,3.0) node{$w_{nm}, b_m$};
%\draw(1,2.75) node{$b_m\in \mathbb{Z}$};
\draw(7,3.0) node{$W_{nm}, B_m$};
%\draw(7.0,3.25) node{$W_{nm}\in \mathbb{R}$};
%\draw(7.0,2.75) node{$B_m\in \mathbb{R}$};
\draw(1,0) node{$s^{out}_m(t)\in \{ 0,1\}$};
\draw(7.0,0) node{$S^{out}_m\in [0, 2^{T-1}]$};
\draw(7.0,6) node{$S^{in}_n\in [0, 2^{T-1}]$};
\draw(1,6) node{$s^{in}_n(t)\in \{ 0,  1 \}$};

\draw[<->, >=stealth] (3,6) -- (5.0,6);
\draw[<->, >=stealth] (3,0) -- (5.0,0);
\draw[<->, >=stealth] (3,3) -- (5.0,3);

\draw[ ->, >=stealth] (7.0,1.98) -- (7.0,0.5);
\draw[->, >=stealth] (1,1.98) -- (1,0.5);
\draw[->, >=stealth] (7.0,5.5) -- (7.0,4);
\draw[->, >=stealth] (1,5.5) -- (1,4);
\draw[->, >=stealth] (0,5.5) -- (0.8,4);
\draw[->, >=stealth] (2,5.5) -- (1.2,4);
\draw[->, >=stealth] (6,5.5) -- (6.8,4);
\draw[->, >=stealth] (8,5.5) -- (7.2,4);
\draw(1,-1) node{\small (a) SNN domain};
\draw(4.1,-1) node{\small (b) Transformation};
\draw(7.2,-1) node{\small (c) ANN domain};
\end{tikzpicture}
\caption{Data transformation and parameter transformation between the SNN domain and the ANN domain.} \label{SNN_ANN}
\end{figure}

\subsection{Fitting into ANN-to-SNN Conversion}

In ANN, the relationship between the input and output can also be expressed as Equation~(\ref{approx}), if we assume $S_n^{in}$ and $S_m^{out}$ to be input and output data, $W_{nm}$ and $B_m$ to be weight and bias. This similarity tells us that the relationship between the input and the output in the SNN domain can be expressed in the ANN domain by transforming the data and the parameters. Table~\ref{transform} shows the transformation methods. WDT and WDT$^{-1}$ are the transformation and inverse transformation functions for input and output data. WPT and WPT$^{-1}$ are the transformation and inverse transformation functions for weights and bias parameters. Note that the input of WDT and the output of WDT$^{-1}$ are time series with index $t$. Figure~\ref{SNN_ANN} clearly shows the relationship between the neuron in the SNN domain and the neuron in the ANN domain. We can describe their relationship in the following two aspects:
\begin{itemize}

\item \textbf{Input/Output} : The input/output data in SNN are time series whose elements are either $0$ or $1$, while the input/output data in ANN are real number ranging from $0$ to $2^{T-1}$. We can transform them by WDT and WDT$^{-1}$.
In traditional ANN with a batch normalization layer, the elements in the input and output feature maps are normalized, which range from $0$ to $1$. Since $T$ is a fixed value, we can scale the data back by $2^T$.

\item \textbf{Weight/Bias}: The weight/bias in SNN is integer while the weight/bias in ANN is a real number. We can transform them by WPT and WPT$^{-1}$. In the traditional ANN with the batch normalization layer, the normalization process is directly applied to the output feature map. Alternatively, we can apply the normalization to trained weights and biases to achieve similar effects~\cite{rueckauer2016theory}.

\end{itemize}

\begin{figure}[!t]
  \centering
  \includegraphics[width=3.5in ]{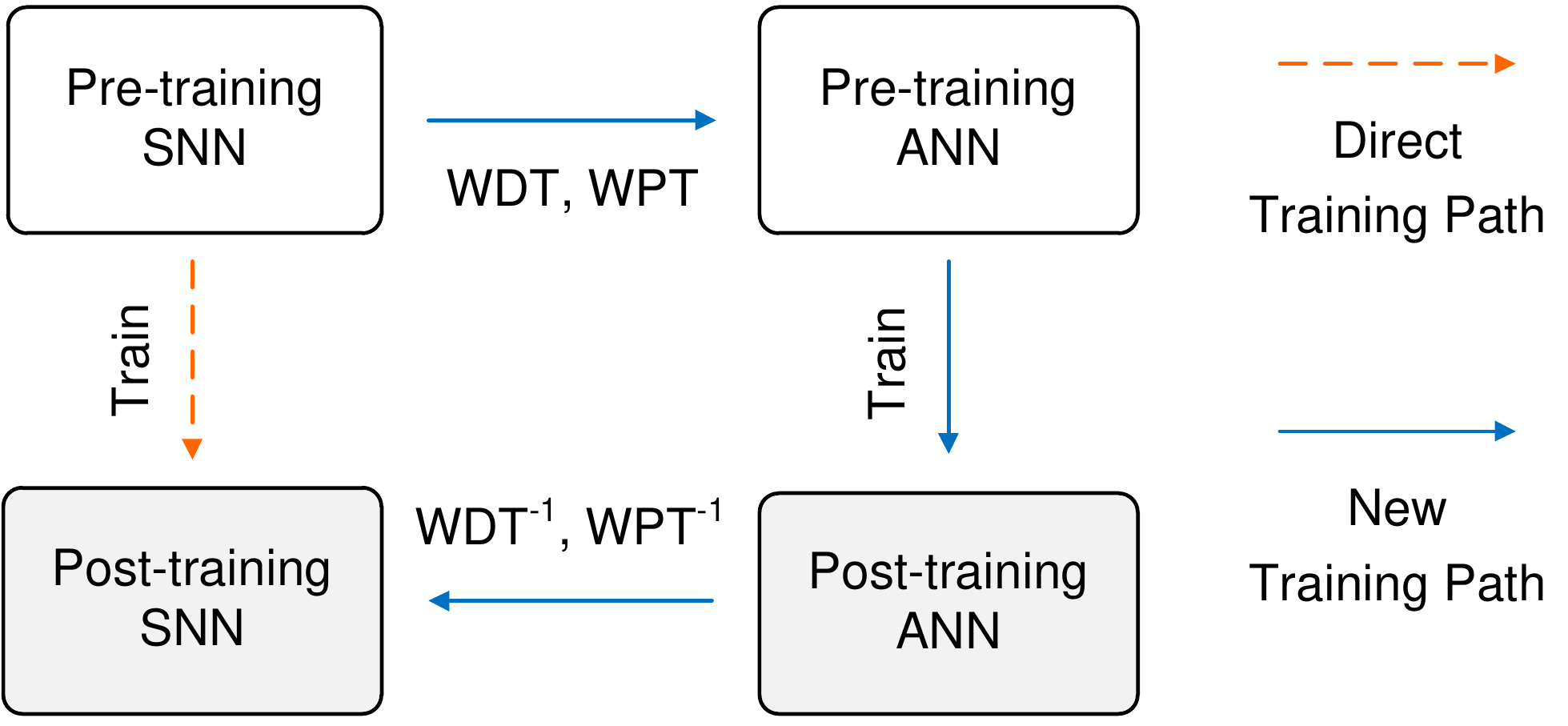}\\
  \caption{The method to fit our radix-encoded SNN into the ANN-to-SNN conversion approach. The dashed red line denotes the direct training path for SNN. The new training path for SNN is shown as the solid blue line.}
  \label{f:traing_path}
\end{figure}

We propose a method to fit our radix-encoded SNN into the ANN-to-SNN conversion approach. Today, training SNN is more difficult than training ANN due to the lack of training platforms and hardware accelerators. The ANN-to-SNN conversion method is proposed~\cite{rueckauer2017conversion} to solve this problem. However, the previous conversion method is based on traditional encoding schemes, and the resulted SNN has low energy efficiency and a larger inference time. By fitting radix encoding into the ANN-to-SNN conversion approach, we make it possible to train radix-encoded SNNs more efficiently on popular platforms and obtain higher accuracy. Figure~\ref{f:traing_path} shows the basic process of our fitting method. In this figure, the dashed red line denotes the direct training path. Following this path, the pre-training SNN model is trained directly to get the post-training SNN model. Alternatively, the new training path is shown as the solid blue line. Following this new path, we first transform the data and parameters from the pre-training SNN model to the pre-training ANN model using the functions WDT and WPT. Then we train the pre-training ANN model to obtain the post-training ANN model using mature platforms and hardware accelerators. Afterward, we can transform the post-training ANN model back to the post-training SNN model using the functions WDT$^{-1}$ and WPT$^{-1}$. Compared with other SNN-to-ANN conversion methods, our approach is more efficient in inference time and energy consumption. We will experimentally prove it in the next section.

\renewcommand{\arraystretch}{1.58}

\begin{table*}[!t]
\caption{Performance of the radix-encoded SNN on the CIFAR-10/100 dataset}
\resizebox{1.00\linewidth}{!}{
\begin{tabular}{l  c   | c  c c c | c  c c  c}
\hline
\hline
&   &  \multicolumn{4}{c|}{CIFAR-10} &\multicolumn{4}{c}{CIFAR-100}\Tstrut\Bstrut\\
Model  & Time Steps & \ Accuracy  & $\Delta$Accuracy & \#Ops ($\times 10^9$)
&Speedup  & Accuracy &$\Delta$Accuracy  & \#Ops ($\times 10^9$)  &Speedup\Bstrut\\

\hline
ResNet-18	&6	&95.26\%	&-0.07\%	&3.34	&167X	&77.44\%	&0.07\%	&3.34	&167X	\Tstrut	\\
	&5	&95.21\%	&-0.12\%	&2.78	&200X	&77.13\%	&-0.24\%	&2.78	&200X~\		\\
	&4	&94.43\%	&-0.90\%	&2.23	&250X	&76.09\%	&-1.28\%	&2.23	&250X	 \Bstrut	\\	
\hline
VGG-16	&6	&93.84\%	&0.08\%	&1.88	&167X	&74.37\%	&0.18\%	&1.88	&167X	\Tstrut	\\
	&5	&93.94\%	&0.18\%	&1.57	&200X	&73.65\%	&-0.54\%	&1.57	&200X~\		\\
	&4	&92.83\%	&-0.93\%	&1.25	&250X	&71.28\%	&-2.91\%	&1.25	&250X	 \Bstrut	\\
\hline
MobileNet	&6	&91.49\%	&0.02\%	&0.28	&167X	&65.50\%	&0.49\%	&0.28	&167X	\Tstrut	\\
	&5	&91.55\%	&0.08\%	&0.23	&200X	&65.11\%	&0.10\%	&0.23	&200X~\		\\
	&4	&91.48\%	&0.01\%	&0.19	&250X	&64.90\%	&-0.11\%	&0.19	&250X	 \Bstrut	\\
\hline
\hline
\label{t:valdiate1}
\end{tabular}
}
\end{table*}

\begin{table*}[!t]
\caption{Performance of the radix-encoded SNN on the ImageNet dataset}
\resizebox{1.00\linewidth}{!}{
\begin{tabular}{l  c   | c  c c c c    c  c}
\hline
\hline
&   &  \multicolumn{7}{c}{ImageNet} \Tstrut\Bstrut\\
Model  & Time Steps &\ \ Top-1 Acc.\ \  &$\Delta$ Top-1 Acc.\ \ & Top-5 Acc.\ \
&$\Delta$ Top-5 Acc.\ \  & \#Ops ($\times 10^9$) &\ Latency \ \   &Speedup\Bstrut\\
\hline
ResNet-18 	&8	&70.24\%	&0.48\%	&89.39\%	&0.31\%	&14.51	&0.008	&125X	\Tstrut	\\
	&6	&69.93\%	&0.17\%	&89.34\%	&0.26\%	&10.88	&0.006	&167X~\		\\
	&4	&66.10\%	&-3.66\%	&86.90\%	&-2.18\%	&7.26	&0.004	&250X	 \Bstrut	\\
\hline
ResNet-34	&8	&73.65\%	&0.35\%	&91.57\%	&0.15\%	&29.31	&0.008	&125X	\Tstrut	\\
	&6	&73.45\%	&0.15\%	&91.51\%	&0.09\%	&21.98	&0.006	&167X~\		\\
	&4	&72.52\%	&-0.78\%	&90.89\%	&-0.53\%	&14.66	&0.004	&250X	 \Bstrut	\\
\hline
\hline
\label{t:valdiate2}
\end{tabular}
}
\end{table*}

\section{Experiment}
\label{sec:experiment}
We implement our radix encoding method in the CUDA-accelerated PyTorch platform. As we take the ANN-to-SNN conversion approach, several popular ANN architectures, including ResNet-18, ResNet-34, VGG-16, and MobileNet are tested. To follow the proposed training path shown in Fig~\ref{f:traing_path}, we implement a customized quantization function and a normalization function. {The weights use fixed-point precision for training.} Each SNN model is trained for several epochs using the straight-through-estimator (STE)~\cite{yin2019understanding} method until the accuracy converges. The SGD optimizer is used with a dynamic learning rate, which decreases its value by half for every fifteen epochs.
{We conducted the experiments on a workstation equipped with Intel i9-10900X CPU, 256 GB memory storage (DRAM), and four Nvidia 2080Ti graphic cards. For training on GPU, we saved the dataset on the SSD (solid-state drive) to save the data loading time. For inference, we simulated the SNN behavior on GPU and manually estimated its latency on the neuromorphic chip. 
{
The estimation is based on a general neuromorphic chip architecture discussed in~\cite{wang2020ncpower}.}
The file system is also compatible with the HDD (hard disk drive) at the cost of a longer training time on GPU, but the estimated inference time on the neuromorphic chip will not be affected.}
For the CIFAR-10 or CIFAR-100 dataset, each experiment can finish within one hour. For the ImageNet dataset, each experiment took around one day.
 
\begin{figure*}[!t]
  \centering
  \includegraphics[width=7in]{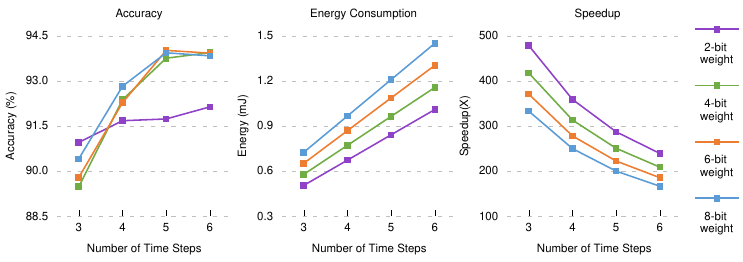}\\
  \caption{Accuracy, energy (based on~\cite{wang2020ncpower}) and speedup of the radix-encoded SNN versus the number of time steps. The speedup is defined as the ratio between the inference time of the rate encoding model over the radix encoding model. Data are collected from the VGG-16 model on the CIFAR-10 dataset.}
  \label{f:Trend}
\end{figure*} 
 
We compare the radix-encoded SNN and rate-encoded SNN models on the neuromorphic chip by counting the number of operations first. The number of operations is calculated based on $\Sigma (N\cdot M)\cdot T$, where $N$ and $M$ are the numbers of input neurons and output neurons in a network layer. $T$ is the number of time steps. We accumulated the number of operations from all network layers. The total number of operations also includes the extra operations for data processing, which reflects the complexity of the encoding scheme. However, even for radix encoded model, only a small number of extra operations are required~\cite{wilamowski2011industrial}. Hence, the total number of operations is roughly proportional to the number of time steps.
The total inference time includes the time to execute all operations plus the extra time for data processing. We assume 
the two encoding schemes use the same amount of hardware resources. The latency of the radix-encoded SNN is normalized by assuming the latency of the rate-encoded SNN to be $1$. We show the speedup of our radix-encoded SNN by comparing its inference time with the traditional rate-encoded SNN. For the rate-encoded SNN, we apply the standard ANN-to-SNN conversion approach~\cite{rueckauer2017conversion} and use a typical configuration and setting~\cite{yan2021near} for near-lossless transition. The number of operations is also roughly proportional to the computational energy consumption. The fewer operations are executed, the less energy is consumed during the computation.

%  For inferece, we simulate the SNN behcanpur on GPU and estaimze its ltaney on the neirophticl chip but it will not be affect by the Choice of file system.

% . For inferece,  we simulate the SNN behcanpur on GPU and estaimze its ltaney on the neuromorphic chip. So the data listed in this paper will not be affected by the location of file system.

% The file system is compatoble to either SSD or HDD<. They show slightlgy differnce in traiing because most of the data set is pre-loaded into the DRAM[]

%  So the latnecy 

% The spike train length $T$ is a fixed parameter. An SNN model can have various choices of $T$. Theoretically, the change of $T$ has non-negligible impacts on the model accuracy and latency. In the experiment part, we will show the performance of radix encoded SNN under different $T$. hard-drive since these can all have an impact on performance. (And would be important for reproducabilityhard-drive since these can all have an impact on performance. 

 \renewcommand{\arraystretch}{1.58}
\begin{table*}[!t]
\caption{Comparison between this work and the state-of-the-art}
\resizebox{1.00\linewidth}{!}{
\begin{tabular}{l  c  c  c   c c c c}
\hline
\hline
Model & Architecture  &  Dataset & Time Steps  & Accuracy \ \  & Acc. Loss  & \#Ops ($\times 10^9$)  & Speedup  \Tstrut \Bstrut\\
\hline
GD-SNN~\cite{sengupta2019going} &VGG-16 & CIFAR-10  &2500 &91.6\%  &-2.2\% &783 & 0.04X\Tstrut\\
%CQ-Training~\cite{yan2021near} &VGG-16 & CIFAR-10  &1000 &93.4\%  &-0.4\% &313  & 0.1X  \\  
Hybrid-SNN~\cite{rathi2020enabling} &VGG-16 & CIFAR-10  &100 &91.1\%  &-2.7\% &31.3  & 1X\\  
Radix Encoding \textbf{(this work)}       &VGG-16   &CIFAR-10 &\textbf{6}  &\textbf{93.8\%} &\ \textbf{0.0\%} &\textbf{1.88}  &\textbf{16.7X}\\ 
Radix Encoding \textbf{(this work)}        &VGG-16   &CIFAR-10 &\textbf{4}  &\textbf{92.8\%} &\textbf{-1.0\%} &\textbf{1.25}  & \textbf{25X}\Bstrut\\ 
\hline
SBP-SNN~\cite{lee2020enabling} &VGG-9   &CIFAR-10 &100  &90.5\% &-2.7\% & 19.1 & 1X\Tstrut\\ 
Radix Encoding \textbf{(this work)}   &VGG-9   &CIFAR-10 &\textbf{4}    &\textbf{91.7\%} &\textbf{-1.5\%} & \textbf{0.76}  & \textbf{25X}\Bstrut\\
\hline
WS-SNN~\cite{kim2018deep} &VGG-20   &CIFAR-10 &200  &90.4\% &-2.0\% & 8.13 & 1X\Tstrut\\ 
Radix Encoding \textbf{(this work)}   &VGG-20   &CIFAR-10 &\textbf{4}    &\textbf{92.2\%} &\textbf{-0.2\%} & \textbf{0.16}  & \textbf{50X}\Bstrut\\
\hline
S-ResNet~\cite{hu2018spiking} &ResNet-44   &CIFAR-100 &350  &68.6\% &-1.6\% & 34.1 & 1X\Tstrut\\ 
Radix Encoding \textbf{(this work)}    &ResNet-44   &CIFAR-100 &\textbf{4}    &\textbf{70.0\%} &\textbf{-0.2\%} & \textbf{0.39}  & \textbf{87.5X}\Bstrut\\
\hline
\hline
\label{t:compare}
\end{tabular}
}
\end{table*}

\subsection{Performance of Radix-Encoded SNN}

To validate our radix encoding method, we test models on the CIFAR-10, CIFAR-100, and ImageNet datasets. In Table~\ref{t:valdiate1} and Table~\ref{t:valdiate2}, we show the inference performance of our radix SNN in terms of accuracy, the number of operations, and the speedup over the rate-encoded SNN. 
{We use the configuration and setting in~\cite{yan2021near} to train baseline rate-encoded SNN models because they can achieve the near-lossless transition from ANN to SNN. The number of time steps is set to be $1000$, a typical value used in~\cite{yan2021near}. Although other methods such as~\cite{sengupta2019going}\cite{rathi2020enabling}\cite{lee2020enabling}\cite{kim2018deep}\cite{hu2018spiking} use a smaller number of time steps, they have a relatively larger accuracy loss. We will compare our method with them in the next part.}
The metrics $\Delta$accuracy stands for the accuracy loss compared to the corresponding rate-encoded SNN model.
From table~\ref{t:valdiate1}, we can see a large speedup with almost no accuracy loss if the radix-encoded technique is applied on SNN. For example, the six-time-step radix-encoded SNN shows around 167X speedup compared to its rate encoding counterpart, on the same accuracy level. This observation is also true for large datasets. From Table~\ref{t:valdiate2}, the radix-encoded SNN using ResNet-18 and ResNet-34 architectures can achieve 167X speedup with even higher model accuracy. As latency is related to the number of time steps, all these advantages of the radix-encoded SNN come from the ultra-short spike trains, where each spike carries extra strength information. This is different from the traditional rate-encoded SNN, where all spikes have the same weight.

Figure~\ref{f:Trend} shows the accuracy, energy consumption, and speedup of radix-encoded SNN versus the number of time steps. Data are collected from the VGG-16 model on the CIFAR-10 dataset. We analyze four types of radix-encoded SNNs, whose weights are quantized by 2 bits, 4 bits, 6 bits, and 8 bits, respectively. The speedup is defined as the ratio between the inference time of the rate encoding model over the radix encoding model. From the figure, we can see that except for the 2-bit weight model, all other models have similar accuracy curves, especially when the number of time steps is greater than five. For the 2-bit weight model, we can still observe over 90\% accuracy. There are two major observations from Figure~\ref{f:Trend}. First, the increment of time steps does not guarantee the increment of accuracy. If the number of time steps is greater than a certain point, the model accuracy has already reached its maximum value. To save energy consumption and lower latency, there is no need to increase the number of steps further. Second, quantizing the weight into more bits does not guarantee the increment of accuracy but causes an increment in energy consumption and latency. In this example, using five time steps and 6-bit weights would be the best choice in terms of accuracy, energy consumption, and latency.

\subsection{Comparison with State-of-the-Art}

We compare our radix-encoded SNN with the state-of-the-art in Table~\ref{t:compare}. The comparison is conducted under the same network architecture and the same dataset. The latencies (speedups) are normalized for easy comparison. From the table, we can see that our work shows not only higher accuracy but also a larger speedup than the state-of-the-art. For example, compared with the Hybrid-SNN~\cite{rathi2020enabling}, the radix-encoded SNN with four time steps can achieve a 25X improvement in latency and 1.7\% improvement in accuracy using the VGG-16 network on the CIFAR-10 dataset. This is mainly due to the reason that the radix encoding method used in the SNN model shows a much lower residual than the rate-encoding method. With only a few time steps, the radix-encoded SNN can achieve very high precision. As we discussed, fewer time steps mean less number of operations. Hence, we can substantially reduce the inference time of the SNN model by the radix encoding method.

\begin{figure}[!t]
  \centering
  \includegraphics[width=3.5in]{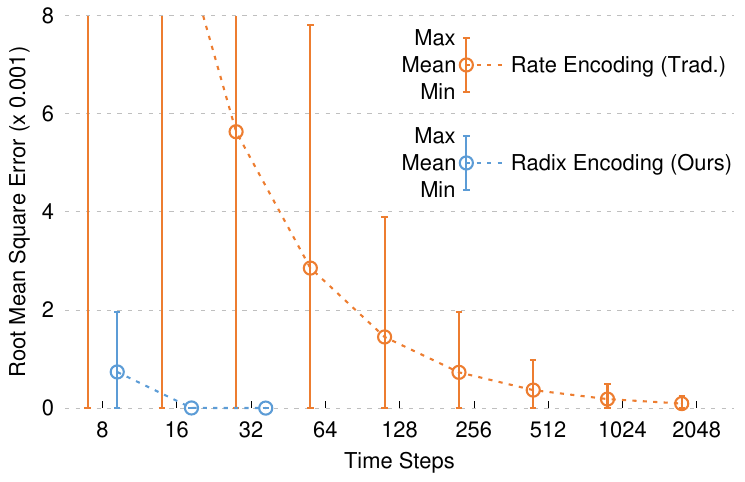}\\
  \caption{Root mean square error between the rate encoding and the radix encoding, the smaller the better. Data are collected from the VGG-16 model on the CIFAR-10 dataset. Activations are normalized to the range [0, 1].}
  \label{f:rmse}
\end{figure}

 \subsection{Root Mean Square Error}

The derivation of the transformation method WDT and WPT is based on the condition that the residual (expressed in Equation~(\ref{residue})) is relatively small and can be ignored. To check the real value of this residual, we evaluate the VGG-16 network on the CIFAR-10 dataset and show the root mean square of the residual under different time steps in Figure~\ref{f:rmse}. In statistics, the root-mean-square of the residual is known as the root mean square error (RMSE)~\cite{chai2014root}. For comparison reasons, we normalize the RMSE by $2^T$, where $T$ is the number of time steps. We study the RMSE in two cases. In the first case, we use the traditional rate encoding method, where each activation is represented by the number of spikes over the spike train length. In the second case, we use our radix encoding method. On the one hand, we want the RMSE value to be as small as possible to achieve the high accuracy of the model. On the other hand, we prefer fewer time steps because it results in fewer operations and smaller latency. 

In Figure~\ref{f:rmse}, we show the RMSE as well as the maximum and minimum residuals. The dashed lines show the trend of the RMSE with the increment of time steps. From the figure, we can see that our radix encoding method shows much less RMSE than that of the traditional rate encoding method. For example, if the number of time steps is eight, the RMSE of the radix encoding method is 96\% less than that of the rate encoding method (this data is not shown in the figure because it is out of the figure's boundary). From another dimension, to achieve the same RMSE, the demanded number of time steps of the radix encoding method is much less than that of the rate encoding method. For example, the radix encoding method with 16 time steps has a similar RMSE as the rate encoding method with 64k time steps.

\section{Conclusion}
\label{sec:conclusion}
In this paper, we proposed an efficient radix-encoded SNN with ultra-short spike trains. We can use less than six time steps to achieve even higher accuracy than its traditional rate-encoded counterpart, which uses one thousand time steps. To train the radix-encoded SNNs more efficiently on mature platforms and hardware, we develop a method to fit the radix encoding technique into the ANN-to-SNN conversion approach. Experiments show that our radix encoding can achieve a 25X improvement in latency and 1.7\% improvement in accuracy compared to the state-of-the-art method using the VGG-16 network on the CIFAR-10 dataset.

\bibliographystyle{IEEEtranN}
\bibliography{ref}

\begin{IEEEbiography}[{\includegraphics[width=1in]{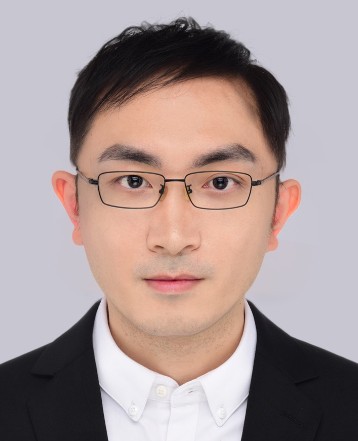}}]{Zhehui Wang}
received the B.S. degree in Electrical Engineering from Fudan University, China, in 2010, and Ph.D. degree in Electronic and Computer Engineering from Hong Kong University of Science and Technology, Hong Kong, in 2016.

He is currently a Research Scientist with the Institute of High Performance Computing, Agency for Science, Technology and Research,
Singapore. His research interests include efficient AI deployment, AI on emerging technologies, hardware-software co-design, and high-performance computing.
\end{IEEEbiography}

\begin{IEEEbiography}[{\includegraphics[width=1in]{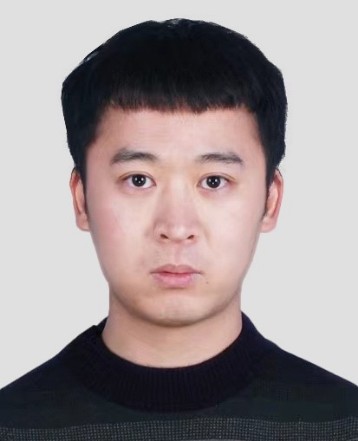}}]{Xiaozhe Gu}
received the Ph.D. degree in Computer Science and Engineering from Nanyang Technological University, Singapore, in 2018. 

He is currently a Post-Doctoral Researcher with the Future Network of Intelligence Institute, the Chinese University of Hong Kong, Shenzhen.  His research interests include real-time system,  scheduling algorithm and machine learning.
\end{IEEEbiography}

\begin{IEEEbiography}[{\includegraphics[width=1in]{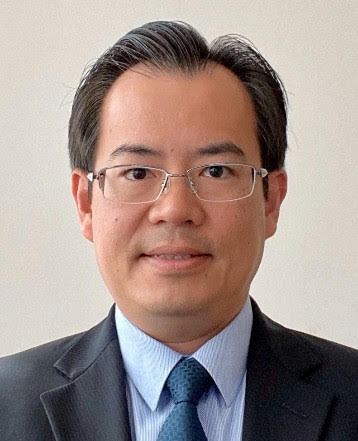}}]{Rick Siow Mong Goh}
received the Ph.D. degree in electrical and computer engineering from the National University of Singapore, Singapore.
He is the Director of the Computing \& Intelligence (CI) Department, Institute of High Performance Computing, Agency for Science, Technology and Research, Singapore, where he leads a team of over 80 scientists in performing world-leading scientific research, developing technology to commercialization, and engaging and collaborating with industry. His current research interests include artificial intelligence, high-performance computing, block chain, and federated learning.
\end{IEEEbiography}

\begin{IEEEbiography}[{\includegraphics[width=1in]{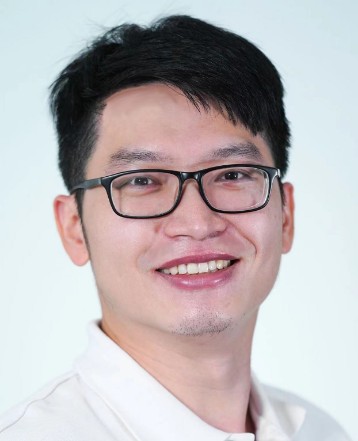}}]{Joey Tianyi Zhou}
is currently a senior scientist, Investigator and group manager with A*STAR Centre for Frontier AI Research (CFAR), Singapore.  He is also holding an adjunct faculty position at National University of Singapore (NUS). Before working at IHPC, he was a senior research engineer with SONY US Research Center in San Jose, USA.  Dr. Zhou received a Ph.D. degree in computer science from Nanyang Technological University (NTU), Singapore. His current interests mainly focus on  machine learning with limited resources and their applications to natural language processing and computer vision tasks. 

Dr. Zhou organized ICDCS'20-21 workshop on Efficient AI meets Edge Computing, ACML'16 workshop on Learning on Big Data workshop and IJCAI'19 workshop on Multi-output Learning; is serving as an Associate Editor for IEEE Transactions on Emerging Topics in Computational Intelligence (TETCI) and IEEE Access, IET Image Processing, and TPC Chair in Mobimedia 2020; and received NeurIPS Best Reviewer Award in 2017.
\end{IEEEbiography}

\begin{IEEEbiography}[{\includegraphics[width=1in]{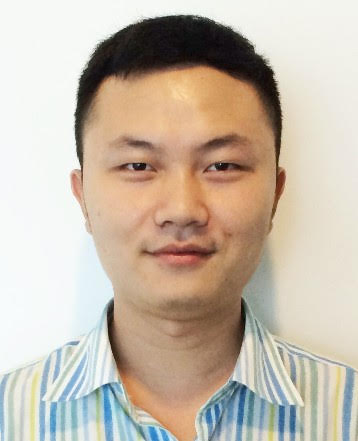}}]{Tao Luo}
received the bachelor’s degree from the Harbin Institute of Technology, Harbin, China, in 2010, the master’s degree from the University of Electronic Science and Technology of China, Chengdu, China, in 2013, and the Ph.D. degree from the School of Computer Science and Engineering, Nanyang Technological University, Singapore, in 2018.

He is currently a Research Scientist with the Institute of High Performance Computing, Agency for Science, Technology and Research, Singapore. His current research interests include high-performance computing, reconfigurable computing system, hardware-software co-exploration, efficient artificial intelligence and its application.
\end{IEEEbiography}

\end{document}